\documentclass[conference]{IEEEtran}
\IEEEoverridecommandlockouts
\usepackage{cite}
\usepackage{amsmath,amssymb,amsfonts,amsthm}
\usepackage{algorithmic}
\usepackage{graphicx}
\usepackage{textcomp}
\usepackage{xcolor}
\usepackage[numbers]{natbib}

\newtheorem{theorem}{Theorem}

\newtheorem{definition}{Definition}
\usepackage[ruled,linesnumbered]{algorithm2e}
\SetKwComment{Comment}{$\triangleright$\ }{}
\usepackage{float}
\usepackage{multirow}
\usepackage{booktabs}

\def\BibTeX{{\rm B\kern-.05em{\sc i\kern-.025em b}\kern-.08em
    T\kern-.1667em\lower.7ex\hbox{E}\kern-.125emX}}

\begin{document}

\title{Decentralized Matrix Factorization with Heterogeneous Differential Privacy

\thanks{Hui Fang$^{*}$ is corresponding author.}
}

\author{\IEEEauthorblockN{Wentao Hu, Hui Fang$^{*}$}
\IEEEauthorblockA{\textit{Research Institute for Interdisciplinary Sciences and School of Information Management and Engineering} \\
\textit{Shanghai University of Finance and
Economics, China} \\
stevenhwt@gmail.com, fang.hui@mail.shufe.edu.cn}
}

\maketitle

\begin{abstract}
Conventional matrix factorization relies on centralized collection of users' data for recommendation, which might introduce an increased risk of privacy leakage especially when the recommender is untrusted. Existing differentially private matrix factorization methods either assume the recommender is trusted, or can only provide a uniform level of privacy protection for all ratings in untrusted recommender scenario.
In this paper, we propose a novel \underline{H}eterogeneous \underline{D}ifferentially \underline{P}rivate \underline{M}atrix \underline{F}actorization algorithm (HDPMF) for untrusted recommender.
To the best of our knowledge, we are the first to achieve heterogeneous differential privacy for decentralized matrix factorization in untrusted recommender scenario. Specifically, our framework uses modified stretching mechanism with an innovative rescaling scheme to achieve better trade-off between privacy and accuracy. Meanwhile, by allocating privacy budget properly, we can capture homogeneous privacy preference within a user/item but heterogeneous privacy preference across different users/items. Theoretical analysis confirms that HDPMF renders rigorous privacy guarantee, and exhaustive experiments demonstrate its superiority especially in strong privacy guarantee, high dimension model and sparse dataset scenario. 
\end{abstract}


\section{Introduction}\label{intro}

Recommender systems have yielded immense success in alleviating information overload and providing personalized services on various online commercial platforms. Despite their effectiveness, they usually need to take full control of users' data to provide accurate recommendations, which may introduce serious privacy leakage since the recommender itself can be untrusted.
An abundance of studies \cite{mcsherry2009differentially,nikolaenko2013privacy} have revealed that an attacker can infer victims' sensitive information such as health conditions \cite{fredrikson2014privacy}, political inclinations and sex inclinations from their ratings on products. The situation gets even worse if the untrusted recommender itself acts as an attacker. Therefore, a privacy-preserving recommender system which can simultaneously provide accurate recommendations and prevent the untrusted recommender from abusing users' ratings is very crucial and necessary in practice.

Among the prevailing collaborative ﬁltering methods for building recommender systems, matrix factorization (MF) \cite{koren2009matrix,candes2009exact} is the most popular and successful one and is being used in many real-world recommender systems, it makes privacy-preserving MF become a research hotspot in recent years. Previous attempts for privacy-preserving MF include data obfuscation \cite{DBLP:conf/nips/XinJ14}, anonymisation \cite{duriakova2019pdmfrec}, differential privacy \cite{DBLP:conf/ijcai/HuaXZ15,DBLP:conf/recsys/LiuWS15} and encryption \cite{nikolaenko2013privacy}. However, most of them suffer from following limitations. First, data obfuscation and anonymisation methods cannot provide a formal privacy guarantee. And the works relying on encryption bring large computation overhead and still cannot escape from inference attacks based on model outputs \cite{mcsherry2009differentially}. Second, a majority of existing works \cite{zhang2019probabilistic,berlioz2015applying,DBLP:conf/recsys/LiuWS15} simply assume the recommender is trusted and neglect the fact that the recommender itself may be untrusted and conduct malicious attacks. Third, most of differentially private MF approaches assume all users and items have the same privacy requirement and thus can only provide uniform privacy protection, which will lead to insufficient protection for some records and over-protection for others, causing dramatic utility loss \cite{jorgensen2015conservative}. A handful of efforts \cite{zhang2019probabilistic,DBLP:conf/recsys/LiuWS15} shed light on personalized differentially private matrix factorization, nevertheless, they only consider trusted recommender scenario, thus cannot handle the situation when recommender is untrusted and acts as an attacker. For example, \cite{zhang2019probabilistic} first samples ratings based on personalized privacy budgets and only use the sampled ratings to conduct MF, others are discarded, then a noise vector is added on the gradient to achieve personalized differential privacy. It is noteworthy that we use personalized differential privacy (PDP) \cite{jorgensen2015conservative} and heterogeneous differential privacy (HDP) \cite{DBLP:journals/jpc/AlagganGK16} interchangeably, they are mathematically equivalent and both mean different ratings have different privacy budgets.

As \cite{zhang2019probabilistic} only use sampled ratings and leaves out unsampled ones to achieve personalized differential privacy in trusted recommender scenario, it aggravates the sparsity problem in recommendation. To avoid this problem, we propose to adopt stretching mechanism \cite{DBLP:journals/jpc/AlagganGK16} in our untrusted recommender scenario, it only needs to stretch ratings and will not discard any of them. First, each user stretches ratings according to personalized privacy budgets on their devices, without disclosing rating values or personalized privacy budgets. Then, each user calculates local gradients with stretched ratings. After adding proper differential privacy noise, the noisy local gradients are sent back to untrusted recommender to update model. Finally, the updated model will satisfy heterogeneous differential privacy. However, if we directly use the original stretching mechanism, the stretching step will distort normal rating scale and harm accuracy, so we remedy it by designing a novel rescaling scheme at prediction stage. In summary, our contributions are as follows:

\begin{enumerate}
\item To the best of our knowledge, we are the first to achieve truly heterogeneous differential privacy for decentralized matrix factorization in untrusted recommender scenario.
\item We tackle the abnormal scale problem \cite{jorgensen2015conservative} in original stretching mechanism \cite{DBLP:journals/jpc/AlagganGK16} with an innovative and efficient rescaling scheme.

\item We allocate privacy budgets for ratings by multiplying privacy preference of the corresponding user and item, thus our method is able to capture homogeneous privacy preference within the same user/item but heterogeneous privacy preference across different users/items.
\item Both theoretical analysis and empirical evaluation verify the effectiveness of our approach, especially in strong privacy guarantee, high dimension model and sparse dataset scenario.

\end{enumerate}

\section{Related Work}
\label{sec:relatedwork}

%
\subsection{Differentially private recommender systems} 

Differential privacy \cite{dwork2006calibrating} has been introduced into recommender systems since \cite{mcsherry2009differentially}, followed by a series of works dedicated on differentially private recommender systems. \cite{zhu2013differential} add noise on item similarity matrix to protect sensitive information in collaborative filtering algorithm and \citep{berlioz2015applying,shin2018privacy,zhang2019probabilistic} add noise on the gradients of matrix factorization algorithm to protect privacy. Nevertheless, a majority of them \cite{berlioz2015applying,shin2018privacy,zhang2019probabilistic} simply posit that recommender is trusted and takes full control of all users' data. \cite{DBLP:conf/ijcai/HuaXZ15} is the first to cope with differentially private recommender system in untrusted recommender scenario. They propose a differentially private matrix factorization approach for untrusted recommender, it decouples computations upon users' sensitive ratings from central recommender to user devices and only allows the untrusted recommender to aggregate local gradients in a privacy-preserving way. However, this method can only provide a uniform privacy guarantee for all ratings. \cite{meng2018personalized} further considers ratings can be divided into sensitive and non-sensitive ones in untrusted recommender scenario. But they assume non-sensitive ratings are public and can only provide a uniform privacy guarantee for sensitive ratings. Therefore, it remains a challenging problem to provide truly personalized differential privacy guarantees for users and items in untrusted recommender scenario.

\subsection{Personalized differential privacy} 
A uniform privacy protection for all records may hamper the performance of differentially private recommenders, since it is common that users have quite personalized privacy expectations meanwhile items have different privacy properties \cite{DBLP:journals/jpc/AlagganGK16}. Consequently, uniform differential privacy may lead to insufficient privacy protection for some records, whilst over-protecting others. \cite{DBLP:journals/jpc/AlagganGK16} is the first work to consider heterogenous privacy preferences of users and items in differential privacy setting. It designs a stretching mechanism that satisfies heterogeneous differential privacy by modifying input values according to their privacy weights and adding noise to final output. However, \cite{jorgensen2015conservative} points out that stretching step in stretching mechanism 
distorts normal scale information and brings significant utility loss. So they propose a novel sample mechanism to achieve personalized differential privacy. It works by first sampling a subset of records based on their personalized privacy budgets. Then, the sampled records will participate in the algorithm and others are left out. Finally, noise is added on the final output to provide personalized privacy guarantee for all records. As part of the ratings are discarded in sample mechanism, it will deteriorate the sparsity problem in recommendation.

\subsection{Federated Matrix Factorization} As one of the most representative decentralized matrix factorization system, federated matrix factorization has attracted much attention in recent years. \cite{ammad2019federated} is the first one to implement federated matrix factorization. They assume it is safe to share gradients without taking any specific privacy protection technique \cite{ammad2019federated,wang2021demystifying}. However, \cite{zhu2019deep,gao2020privacy} point out that original data can be leaked from unprotected gradients in federated learning. To prevent the leakage from shared gradients in federated matrix factorization, an abundance of work rely on cryptography techniques such as homomorphic encryption \cite{chai2020secure,du2021federated} and secure aggregation \cite{li2021federated,wan2022towards}. Nevertheless, they bring huge amount of computation overhead and still cannot escape the vulnerability to inference attacks \cite{dinur2003revealing,li2021federated}. On the other hand, a handful of works borrow the idea of meta learning to implement federated matrix factorization \cite{lin2020meta,singhal2021federated}. They do not add any noise and cannot provide provable privacy guarantee, so it is unfair to compare our HDPMF with them, as HDPMF involves adding noise and providing rigorous privacy guarantee.

Combining the promising characteristics of provable resistance to inference attacks and rigorous privacy guarantee, differential privacy (DP) has been used to protect federated learning \cite{li2021federated,liu2021projected}.
For example, \cite{li2021federated} provide rigorous privacy guarantee for federated matrix factorization with the help of DP, they combine DP with secure aggregation to provide a uniform privacy guarantee across different users.
On the contrary, our HDPMF framework can provide personalized privacy guarantees for different users in such decentralized matrix factorization scenario.



\section{Preliminaries}
\label{sec:preliminaries}
\subsection{Differential Privacy}
Differential privacy \cite{dwork2006differential} ensures that the output of any randomized algorithm is insensitive to the change of a single record by injecting proper noise to intermediate process or final output. We formally give its definition as below:

\begin{definition}{\textbf{$\epsilon$-differential privacy.}}
A randomized function $f:\mathbb{R}^{n}\rightarrow\mathbb{R}$ is said to be $\epsilon$-differentially private, if for all neighboring datasets that differ only on a single record $i$, denoted as $\vec{d}\sim\vec{d}^{(i)} \in \mathbb{R}^{n}$, and for all outputs $t \in \mathbb{R}$, the following statement holds:
\begin{equation}
\operatorname{Pr}[f(\vec{d})=t] \leq \exp (\epsilon) \operatorname{Pr}[f(\vec{d}^{(i)})=t]
\end{equation}
\emph{where the probability is taken over the randomness of $f$. Privacy budget $\epsilon$ is a positive number, and a smaller $\epsilon$ means a stronger privacy guarantee.}
\end{definition}

\emph{Laplace mechanism} \cite{dwork2006calibrating} is the most commonly used method to satisfy $\epsilon$-differential privacy. It works by adding i.i.d. noise from Laplace distribution with location parameter $0$ and scale parameter $S(f)/\epsilon$, denoted as $Laplace(S(f)/\epsilon)$. Here, global sensitivity $S(f)=\max_{\vec{d}, \vec{d}^{(i)}}\|f(\vec{d})-f(\vec{d}^{(i)})\|_1$ is the maximum change of output caused by any single record $i$ in the input $\vec{d}$.

\begin{definition}{\textbf{heterogeneous $(\epsilon, \vec{v})$-differential privacy.}}  \label{def_hdp}
    For all semi-balanced sets $D$, a randomized function $f: D \rightarrow \mathbb{R}$ is said to be $(\epsilon, \vec{v})$-differentially private if for all records $i$ and all neighboring datasets $\vec{d} \sim\vec{d}^{(i)}\in D$, and for all outputs $t \in \mathbb{R}$, the following statement holds:
    \begin{equation}
      \operatorname{Pr}[f(\vec{d})=t] \leq \exp (\epsilon v_{i}) \operatorname{Pr}[f(\vec{d}^{(i)})=t]  
    \end{equation}
\emph{where semi-balanced set $D$ means that for any $\vec{x} \in D$, then $A\vec{x} \in D$, where $A$ is a diagonal shrinkage matrix whose diagonal elements are in the range of $(0,1]$. Rating matrix can be flattened into a non-negative vector, then it is apparently in a semi-balanced set, so we aim to achieve heterogeneous differential privacy for rating matrix in our work. $\vec{v} \in (0,1]^n$ is the privacy weight vector associated with records in dataset $\vec{d}$ and $\epsilon$ is the maximum privacy budget of all records.} 
\end{definition}

\subsection{Matrix Factorization}
We consider the standard matrix factorization model with regularization term \cite{koren2009matrix}. $\boldsymbol{R} \in \mathbb{R}^{n \times m}$ is the full rating matrix, in which each of
$n$ users rates a small subset of $m$ items. We denote user $i$'s rating on item $j$ as $\boldsymbol{R}_{ij}$ and $\boldsymbol{I}$ is a indicator matrix that $\boldsymbol{I}_{ij}=1$ when user $i$ has rated item $j$, otherwise  $\boldsymbol{I}_{ij}=0$. We can decompose rating matrix $\boldsymbol{R}$ into user latent matrix $\boldsymbol{U}=[\boldsymbol{u}_i]_{i \in [1,n]} \in \mathbb{R}^{K \times n}$ and item latent matrix $\boldsymbol{V}=[\boldsymbol{v}_j]_{j \in [1,m]} \in \mathbb{R}^{K \times m}$, where each user $i$ is characterized by a user latent vector $\boldsymbol{u}_i \in \mathbb{R}^K$ and each item $j$ is characterized by a item latent vector $\boldsymbol{v}_j \in \mathbb{R}^K$, $K$ is the dimension of all latent vectors. Finally, the matrix factorization model can be expressed as minimizing the loss function $\mathcal{L}$:
\begin{equation*}
  \label{nonp_obj}
    \min _{\boldsymbol{U}, \boldsymbol{V}} \mathcal{L} =\sum_{i=1}^{n} \sum_{j=1}^{m} \boldsymbol{I}_{i j}(\boldsymbol{R}_{ij}-\boldsymbol{u}_{i}^{T} \boldsymbol{v}_{j} )^{2}  +\lambda(||\boldsymbol{U}\|_{F}^{2}+\|\boldsymbol{V}\|_{F}^{2})
\end{equation*}
where $||\cdot||_F^2$ is the Frobenius norm and $\lambda$ is regularization coefficient. We use stochastic gradient descent with learning rate $\eta$ to minimize the objective function.

\subsection{Problem Statement}
In a basic recommender system, there are two types of roles: users and the recommender. We consider the recommender is untrusted, which means that it is curious about users' sensitive ratings and even misuses them for profits. And, all ratings are considered heterogeneously sensitive, that is, when user $i$ rates item $j$ (i.e. $\boldsymbol{I}_{ij}$=1), it will also specify a non-public privacy weight $\boldsymbol{W}_{ij} \in (0,1]$ for rating $\boldsymbol{R}_{ij}$ according to user $i$'s own privacy preference and privacy property of item $j$. That is to say, the personalized privacy budget of rating $\boldsymbol{R}_{ij}$ is $\epsilon_{ij}= \boldsymbol{W}_{ij} \cdot \epsilon$, where $\epsilon$ is the maximum privacy budget allowed for all ratings.

Our objective is to design a differentially private matrix factorization framework that satisfies heterogeneous differential privacy. As the recommender is untrusted, it can only access to the item latent matrix $\boldsymbol{V}$, whereas user latent matrix $\boldsymbol{U}$ and all ratings $\boldsymbol{R}_{ij}$ are kept on users' devices privately. When updating model parameters, the recommender needs to distribute item latent matrix $\boldsymbol{V}$ to users' devices and collects noisy gradients of $\boldsymbol{V}$ sent from users. On user side, users need to calculate local gradients and send them to recommender after adding noise. At prediction stage, user $i$ can make predictions privately on its device with its own latent vector $\boldsymbol{u}_i$ and the shared item latent vector $\boldsymbol{V}$.

\section{Our HDPMF Approach}
\label{sec:approach}

As sample mechanism in \cite{zhang2019probabilistic} will deteriorate the sparsity problem in recommendation, we propose to integrate \emph{stretching mechanism} \cite{DBLP:journals/jpc/AlagganGK16} with objective perturbation \cite{chaudhuri2011differentially} to achieve heterogeneous differential privacy. The key component of \emph{stretching mechanism} is to find a proper diagonal shrinkage matrix $T(\vec{v})$ with all diagonal values in $(0,1]$ based on its input $\vec{d}$ and corresponding privacy weight vector $\vec{v} \in (0,1]^n$. Suppose we want to make a function $f$ satisfy heterogeneous differential privacy, after getting the proper shrinkage matrix $T(\vec{v})$, we then need to modify the input values $\vec{d}$ by multiplying it with $T(\vec{v})$. Thus, the modified computation can be denoted as $R(f,\vec{v})(\vec{d})=f(T(\vec{v}) \cdot \vec{d})$. After that, the \emph{stretching mechanism} can be expressed as: $\mathcal{SM}(\vec{d},\vec{v},\epsilon)=R(f,\vec{v})(\vec{d})+Laplace(S(f)/\epsilon)$, where $S(f)$ is global sensitivity of $f$. As heterogeneous differential privacy also involves the exact influence of the change of record $i$ in $\vec{d}$ on the final output, we then rely on the notion of \emph{modular global sensitivity} \cite{dandekar2011privacy} to denote it.

\begin{definition}{\textbf{modular global sensitivity.}} \label{modular}
        The modular global sensitivity $S_{i}(f)$ is the global sensitivity of $f$ when $\vec{d}$ and $\vec{d}^{(i)}$ are neighboring datasets that differ on exactly the record $i$.
\end{definition}
With the help of \emph{modular global sensitivity}, Theorem \ref{stretch} guarantees that \emph{stretching mechanism} satisfies heterogeneous differential privacy in Definition \ref{def_hdp}.

\begin{theorem} \label{stretch}
    Given a privacy weight vector $\vec{v}$, if the shrinkage matrix $T(\vec{v})$ satisfies $S_{i}(R(f, \vec{v})) \leq v_{i} S(f)$, where $R(f,\vec{v})(\vec{d})=f(T(\vec{v}) \cdot 
     \vec{d})$, then the stretching mechanism $\mathcal{S M}(\vec{d}, \vec{v}, \epsilon)=R(f, \vec{v})(\vec{d})+ Laplace(S(f)/\epsilon)$ satisfies heterogeneous $(\epsilon, \vec{v})$-differential privacy.
\end{theorem}

\begin{proof}
Please see the proof of Theorem 1 in \cite{DBLP:journals/jpc/AlagganGK16}.
\end{proof}

As the gradient $\frac{\partial \mathcal{L}}{\partial \boldsymbol{u}_i}$ only involves $\boldsymbol{u}_i$ and $\boldsymbol{V}$, each user $i$ can update its $\boldsymbol{u}_i$ locally on its device privately. Therefore, as long as item latent matrix $\boldsymbol{V}$ satisfies differential privacy, user latent vector $\boldsymbol{u}_i$ derived from it will satisfy differential privacy as well \cite{DBLP:conf/ijcai/HuaXZ15}. Similarly, as heterogeneous differential privacy has the nice property of post-processing \cite{DBLP:journals/jpc/AlagganGK16}, if item latent matrix $\boldsymbol{V}$ satisfies heterogeneous $(\epsilon,\vec{v})$-differential privacy, then correspondingly the user latent vectors derived from it will also satisfy heterogeneous $(\epsilon,\vec{v})$-differential privacy.

In HDPMF framework,  we first flatten the rating matrix $\boldsymbol{R}$ into a input vector $\vec{d}$ and flatten the privacy weight matrix $\boldsymbol{W}$ into the privacy weight vector $\vec{v}$. Meanwhile, we let the shrinkage matrix $T(\vec{v})=\operatorname{diag}(\vec{v})$, which means that we modify the input rating matrix $\boldsymbol{R}$ through element-wise product with $\boldsymbol{W}$. After that, some calibrated Laplace noise is added to satisfy heterogeneous differential privacy. The whole process can be formulated as follows:  
\begin{equation}
  \label{priv_obj}
  \begin{split}
    \min _{\boldsymbol{U}, \boldsymbol{V}} \widetilde{\mathcal{L}} &=\sum_{i=1}^{n} \sum_{j=1}^{m} \boldsymbol{I}_{i j}((\boldsymbol{W}_{ij}\boldsymbol{R}_{ij}-\boldsymbol{u}_{i}^{T} \boldsymbol{v}_{j} )^{2}+\boldsymbol{v}_j^T \boldsymbol{x}_j^i) \\ &+\lambda(||\boldsymbol{U}\|_{F}^{2}+\|\boldsymbol{V}\|_{F}^{2})
    \end{split}
\end{equation}
 where $\boldsymbol{x}_j=\sum_{i}\boldsymbol{x}_j^i \in \mathbb{R}^K $ denotes noise vector.
 
 We can derive the gradients of Equation (\ref{priv_obj}) as follows:
\begin{equation}
    \label{priv_grad}
    \begin{split}
        &\frac{\partial \widetilde{\mathcal{L}}}{\partial \boldsymbol{v}_{j}} =2\sum_{i=1}^{n} \boldsymbol{I}_{i j}(\boldsymbol{u}_{i}^{T} \boldsymbol{v}_{j}-\boldsymbol{W}_{ij}\boldsymbol{R}_{i j}) \boldsymbol{u}_{i} +\boldsymbol{x}_j+2 \lambda \boldsymbol{v}_{j} \\
        &\frac{\partial \widetilde{\mathcal{L}}}{\partial \boldsymbol{u}_{i}} =2\sum_{j=1}^{m} \boldsymbol{I}_{i j}(\boldsymbol{u}_{i}^{T} \boldsymbol{v}_{j}-\boldsymbol{W}_{ij}\boldsymbol{R}_{i j}) \boldsymbol{v}_{j} +2 \lambda \boldsymbol{u}_{i}  
    \end{split}
\end{equation}

\begin{theorem}
\label{th2}
  If each element in $\boldsymbol{x}_j$ is independently and randomly sampled from $Laplace(2\sqrt{K}\Delta/\epsilon)$, where $\Delta=\max \boldsymbol{R}_{ij}-\min\boldsymbol{R}_{ij}$, then the derived item latent  matrix $\boldsymbol{V}$  satisfies heterogeneous $(\epsilon, \vec{v})$-differential privacy.
\end{theorem}
\begin{proof}
Please see Appendix.
\end{proof}

Theorem \ref{th2} shows that the final learned item latent matrix $\boldsymbol{V}$ from Equation (\ref{priv_obj}) can satisfy heterogeneous $(\epsilon,\vec{v})$-differential privacy. The only problem left is how can users independently pick $\boldsymbol{x}_j^i$ and make sure that $\boldsymbol{x}_j=\sum_{i}\boldsymbol{x}_j^i \sim Laplace(2\sqrt{K}\Delta/\epsilon)$. We can utilize Theorem \ref{composition} to achieve this goal. The recommender first picks $\boldsymbol{h}_j \in \mathbb{R}^K \sim Exp(1)$ and distributes it to the set of users $\mathcal{R}_j$ who have rated item $j$. Then, each user in $\mathcal{R}_j$ independently selects $\boldsymbol{c}_{j}^i \in \mathbb{R}^K \sim N(0,1/|\mathcal{R}_j|)$ and takes $\boldsymbol{x}_j^i=2\Delta\sqrt{2K \boldsymbol{h}_j} \odot \boldsymbol{c}_j^i/\epsilon$, where $\odot$ means element-wise product. According to Theorem \ref{composition}, $\sum_{i \in \mathcal{R}_j}\boldsymbol{x}_j^i \sim Lap(2\Delta \sqrt{K}/\epsilon)$.

\begin{theorem}
\label{composition}
If random number $h \sim Exp(1)$, and random number $c \sim N(0,1)$ is independent of $h$, then for any real number $b >0$, $X=b \sqrt{2 h} c \sim Laplace(b)$ \cite{kotz2001laplace}. 
\end{theorem}

\begin{figure}[htb!]
    \centering
    \includegraphics[scale=0.26]{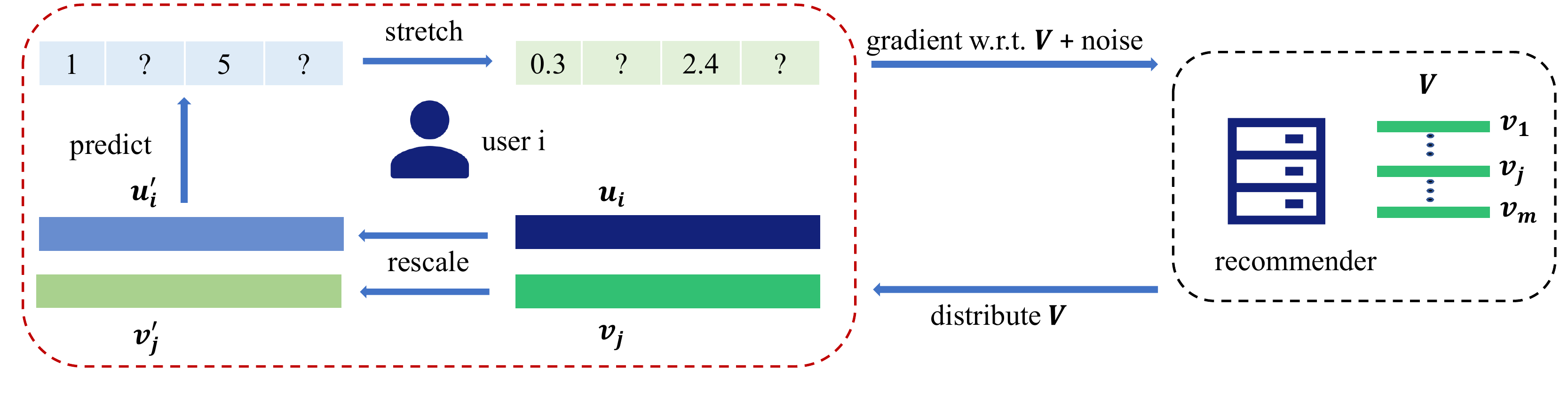}
    \caption{The framework of HDPMF.} 
    \label{hdpmf_framework}
\end{figure}

The framework of HDPMF is depicted in Figure \ref{hdpmf_framework} and its pseudo-code is shown in Algorithm \ref{alg1}. Users first stretch ratings according to personalized privacy weights on their devices. Then, they use stretched ratings to calculate local gradients with respect to $\boldsymbol{V}$. After adding noise $\boldsymbol{x}_j^i$, noisy local gradients are sent to the untrusted recommender. With the collected noisy local gradients sent from users, the untrusted recommender updates item latent matrix $\boldsymbol{V}$ and distributes it to users. After that, users update their own latent vector $\boldsymbol{u}_i$ privately on their devices with shared item latent matrix $\boldsymbol{V}$.


As \cite{jorgensen2015conservative} points out that stretching the rating matrix distorts the normal scale information and causes significant utility loss, so we design an innovative rescaling scheme at prediction stage to mitigate the negative effect brought by stretching procedure. The rescaling scheme runs on on users' devices privately, so it will not breach users' privacy and  it is shown in lines 13-17 in Algorithm \ref{alg1}. In rescaling scheme, if user $i$ wants to predict its preference on item $j$, it must first rescale $\boldsymbol{v}_j$ and $\boldsymbol{u}_i$ based on personalized privacy weight $\boldsymbol{W}_{ij}$. After that, it multiplies recaled $\boldsymbol{v}_j$ and $\boldsymbol{u}_i$ to get normal scale prediction. We will empirically demonstrate the effectiveness of rescaling scheme for reducing utility loss in the ablation study.

\begin{algorithm}[tb]
        \caption{HDPMF}\label{alg1}
        \SetAlgoNoLine
        \DontPrintSemicolon
        \KwIn{$\boldsymbol{R}$, $\boldsymbol{W}$, $\epsilon$, $K$, $\eta$, $\lambda$, $T$}
        \KwOut{predicted rating matrix $\widehat{\boldsymbol{R}}$}
        \SetKw{KwFrom}{from}
        \SetKw{KwInit}{initialize}
        \KwInit  $\boldsymbol{U}, \boldsymbol{V}$  \\
             \For{t \KwFrom 1 \KwTo T}{
                \For{$j=1, \cdots,m$ }{
                  \For{i in $\mathcal{R}_j$}{
                        Send 2$(\boldsymbol{u}_{i}^{T} \boldsymbol{v}_{j}-\boldsymbol{W}_{ij}\boldsymbol{R}_{i j}) \boldsymbol{u}_{i}+\boldsymbol{x}_j^i$ to the recommender
                  }
                   \Comment*[l]{on recommender}
                  Update $\boldsymbol{v}_j$:   $\boldsymbol{v}_j=\boldsymbol{v}_j-\eta \frac{\partial \widetilde{\mathcal{L}}}{\partial \boldsymbol{v}_{j}}$
                 }
                 \For{$i=1, \cdots,n$}{
                 \Comment*[l]{on user $i$'s device}
                  Update $\boldsymbol{u}_i$:  $\boldsymbol{u}_i=\boldsymbol{u}_i-\eta \frac{\partial \widetilde{\mathcal{L}}}{\partial \boldsymbol{u}_{i}}$
                 }
             }
             \For{$i=1, \cdots,n$}{
            \Comment*[l]{rescale on user $i$'s device} 
              \For{$j=1, \cdots,m$}{
            $\widehat{\boldsymbol{R}}_{ij}=\boldsymbol{u}_i^T \boldsymbol{v_j}/\boldsymbol{W}_{ij}$
            }
            }
             \Return $\widehat{\boldsymbol{R}}$
\end{algorithm}

\section{Experimental Evaluation}
In this section, we conduct extensive experiments to validate the effectiveness of HDPMF framework. 

\begin{table}[tb!]
  \centering
  \caption{Statistics of datasets after preprocessing}
    \begin{tabular}{lllll}
    \hline
    Dataset & \#users & \#items & \#ratings & \#density \\
    \hline
    ML-100K & 1,682  & 943   & 100,000 & 4.26e-2 \\
    ML-1M & 6,040  & 3,883  & 1,000,209 & 6.30e-2 \\
    Douban & 1,403 & 94,651 & 221,195 & 1.67e-3 \\
    Yelp & 36,997 & 147,435 & 1,888,380 & 3.46e-3 \\
    \hline
    \end{tabular}%
    \label{statistics}
\end{table}%

\subsection{Experimental Settings}
\subsubsection{Evaluation Metrics.}
We divide datasets into training and validation sets using $5$-fold cross validation for hyperparameter selection. Following ML-100K, at test stage, each user leaves $10$ ratings as test dataset, and the rest ratings as training dataset. As we focus on the rating prediction problem, we adopt mean square error (MSE) and mean absolute error (MAE) as evaluation metrics, which are defined as $\sum_{(i,j) \in \mathcal{R}} (\widehat{\boldsymbol{R}}_{ij}-\boldsymbol{R}_{ij})^2/|\mathcal{R}|$ and $\sum_{(i,j) \in \mathcal{R}} |\widehat{\boldsymbol{R}}_{ij}-\boldsymbol{R}_{ij}|/|\mathcal{R}|$ respectively, $\mathcal{R}$ denotes ratings in test dataset. Each evaluation is repeated $5$ times using different random seeds.

\subsubsection{Baselines and hyperparameters.} We compare HDPMF with state-of-the-art differentially private matrix factorization methods and non-private method.

\textbf{PDPMF}: \cite{zhang2019probabilistic} proposes a personalized differentially private matrix factorization method in trusted and centralized recommender. As it's for centralized recommender, it only adds a single centralized noise vector on the gradient. We modify this approach with Theorem \ref{composition} to decompose the single noise vector to many local noise vectors, so we can use these local noise vectors to protect each user's local gradient in untrusted and decentralized recommender.

\textbf{DPMF}: a decentralized and non-personalized differentially private matrix factorization method that adds equal noise for all ratings \cite{DBLP:conf/ijcai/HuaXZ15}.

\textbf{MF}: a non-private and centralized matrix factorization method that decomposes rating matrix without adding any noise.




For all the methods, regularization coefficient $\lambda$ is picked from $\{0.01,0.001\}$. The learning rate $\eta$ stays the same as initial learning rate in the first $25\%$ of the epochs and is divided by $5$ after that, then is divided by $5$ again after $75\%$ of the epochs, which is similar to the decay scheme in \cite{abadi2016deep}. We test initial learning rate from $\{0.05,0.01,0.005,0.001\}$ for HDPMF, PDPMF, MF and $\{0.005,0.001,0.0005,0.0001\}$ for DPMF. The model dimension $K$ is 10 in default setting following \cite{meng2018personalized}, we also test $K=5$. The number of epochs is set as $T=100$ for all experiments.

\subsubsection{Privacy Speciﬁcation.} To capture homogeneous privacy preference within the same user/item but heterogeneous privacy preference across different users/items. We assume each user $i$ has a personalized privacy weight $\beta_i$, then the user privacy weight vector $\boldsymbol{\beta}=(\beta_1,\beta_2, \cdots, \beta_n)$. Similarly, each item $j$ has a privacy weight $\gamma_j$ and item privacy weight vector can be expressed as $\boldsymbol{\gamma}=(\gamma_1,\gamma_2, \cdots, \gamma_m)$. Finally, we let the privacy weight matrix $\boldsymbol{W}=\boldsymbol{\beta}^T \boldsymbol{\gamma}$.   

We follow \cite{jorgensen2015conservative,zhang2019probabilistic} to use random allocation to simulate personalized privacy specification setting.  We set the maximum privacy budget $\epsilon=1$. And users are divided into 3 groups: \emph{conservative}, \emph{moderate} and \emph{liberal} whose privacy concern is high, medium, low respectively and their corresponding ratios are $f_{uc}$, $f_{mc}$ and $1-f_{uc}-f_{mc}$, respectively. Similarly, items can also be divided into 3 groups which are highly, moderately and least sensitive with corresponding ratio $f_{ic},f_{im}$ and $1-f_{ic}-f_{im}$. The privacy weight for the users in \emph{conservative} and \emph{moderate} groups are randomly and uniformly drawn from the ranges $[\epsilon_{uc},\epsilon_{um})$ and $[\epsilon_{um},\epsilon_{ul})$, whilst the privacy weight is fixed at $\epsilon_{ul}=1$ for \emph{liberal} group.  As for the privacy weights of items, we draw uniformly at random from $[\epsilon_{ic},\epsilon_{im})$ and $[\epsilon_{im},\epsilon_{il})$ for highly and moderately sensitive items. Similarly, privacy weights are fixed at $\epsilon_{il}=1$ for least sensitive items. The overview of personalized privacy specification parameters and their default values are shown in Table \ref{privacy}.



  

\begin{table}[htbp]
  \centering
  \caption{Privacy specification parameters and their default values. con, mod, lib mean conservative, moderate and liberal.}
    \begin{tabular}{llllll}
    \hline
     & group & ratio & default & weight & default \\
    \hline
    \multirow{3}[2]{*}{user} & con & $f_{uc}$   & $0.54$  & $[\epsilon_{uc},\epsilon_{um})$     & $[0.1,0.5)$ \\
          & mod  & $f_{um}$   & $0.37$  & $[\epsilon_{um},\epsilon_{ul})$     & $[0.5,1)$ \\
          & lib & $1$-$f_{uc}$-$f_{um}$ & $0.09$  & $\epsilon_{ul}$     & $1$\\
    \hline
    \multirow{3}[2]{*}{item} & con & $f_{ic}$   & $0.33$  & $[\epsilon_{ic},\epsilon_{im})$     & $[0.1,0.5)$ \\
          & mod  & $f_{im}$   & $0.33$  & $[\epsilon_{im},\epsilon_{il})$     & $[0.5,1)$ \\
          & lib & $1$-$f_{ic}$-$f_{im}$ & $0.34$  & $\epsilon_{il}$     & $1$ \\
    \hline
    \end{tabular}%
  \label{privacy}%
\end{table}%

\begin{table*}[htb!]
  \centering
  \caption{Average and standard deviation of MSE and MAE on 5 repeated evaluations at the default privacy specification setting. Bold number means HDPMF outperforms PDPMF on the average of MSE or MAE. $^{***}$ means HDPMF outperforms PDPMF at 99\% confidence level in t-test for means of paired two sample and $^{**}$ means 95\%, $^{*}$ means 90\% confidence level correspondingly.}
    \begin{tabular}{c|c|llll}
    \hline
    \multicolumn{2}{c}{metrics} & \multicolumn{2}{c}{MSE} & \multicolumn{2}{c}{MAE} \\
    \cmidrule(lr){3-4}  \cmidrule(lr){5-6}  
    \multicolumn{2}{c}{dimension} & \multicolumn{1}{c}{K=10} & \multicolumn{1}{c}{K=5} & \multicolumn{1}{c}{K=10} & \multicolumn{1}{c}{K=5} \\
    \hline
    \multirow{4}[2]{*}{ML-100K} 
    & MF  & 0.9269$\pm$ 0.0042   & 0.9231$\pm$ 0.0065  & 0.7617 $\pm$ 0.0015 & 0.7609 $\pm$ 0.0031     \\
          & HDPMF & \textbf{1.4690 $\pm$ 0.0369$^{***}$} & \textbf{1.2257 $\pm$ 0.0187$^{*}$} & \textbf{0.9356 $\pm$ 0.0124$^{***}$} & \textbf{0.8606 $\pm$ 0.0045$^{***}$}   \\
          & PDPMF & 1.5623 $\pm$ 0.0244 & 1.2475 $\pm$ 0.0214 & 0.9664 $\pm$ 0.0057 & 0.8715 $\pm$ 0.0078   \\
          & DPMF  & 4.9264 $\pm$ 0.3218 & 4.4484  $\pm$ 0.2955 & 1.8811 $\pm$ 0.0783 & 1.7685 $\pm$ 0.0719  \\
    \hline
    \multirow{4}[2]{*}{ML-1M} 
    & MF  & 0.7906 $\pm$ 0.0012  & 0.8067 $\pm$ 0.0021  & 0.7027 $\pm$ 0.0007 & 0.7104 $\pm$ 0.0010   \\
          & HDPMF & \textbf{1.0262 $\pm$ 0.0057$^{***}$}  & \textbf{0.9174 $\pm$ 0.0063$^{***}$} & \textbf{0.7869 $\pm$ 0.0028 $^{***}$} & \textbf{0.7501 $\pm$ 0.0023$^{***}$}  \\
          & PDPMF & 1.0801 $\pm$ 0.0192 & 0.9399 $\pm$ 0.0060 & 0.8136 $\pm$ 0.0065 & 0.7635 $\pm$ 0.0018  \\
          & DPMF  & 1.8032 $\pm$0.0271  & 1.3912 $\pm$ 0.0114 & 1.0350 $\pm$ 0.0078 & 0.9166 $\pm$ 0.0035  \\
    \hline
     \multirow{4}[2]{*}{Douban} 
    & MF  & 1.5403 $\pm$ 0.0057   & 1.5143  $\pm$ 0.0079  & 0.9569 $\pm$ 0.0017 & 0.9481 $\pm$ 0.0020   \\
          & HDPMF & \textbf{4.5616  $\pm$ 0.1013$^{***}$}  & \textbf{3.6169  $\pm$ 0.0843$^{***}$} & \textbf{1.7451  $\pm$ 0.0296  $^{***}$} & \textbf{1.4929  $\pm$ 0.0232 $^{***}$}  \\
          & PDPMF & 5.0464 $\pm$ 0.1215 & 4.2775  $\pm$ 0.0676  & 1.8677  $\pm$ 0.0277  & 1.6786  $\pm$ 0.0155   \\
          & DPMF  & 5.9369 $\pm$0.0574   & 6.1115 $\pm$ 0.0614 & 2.0250  $\pm$ 0.0138  & 2.0746  $\pm$ 0.0154   \\
    \hline
     \multirow{4}[2]{*}{Yelp} 
    & MF  & 1.9080 $\pm$ 0.0098   & 1.9141  $\pm$ 0.0317  & 1.0795 $\pm$ 0.0030 & 1.0784 $\pm$ 0.0081   \\
          & HDPMF & \textbf{3.5206   $\pm$ 0.0184$^{***}$}  & \textbf{2.8944  $\pm$ 0.0220$^{**}$} & \textbf{1.4661   $\pm$ 0.0052   $^{***}$} & \textbf{1.3062   $\pm$ 0.0055  $^{***}$}  \\
          & PDPMF & 3.8329 $\pm$ 0.0341 & 2.9528  $\pm$ 0.0324  & 1.5351   $\pm$ 0.0068   & 1.3325   $\pm$ 0.0079    \\
          & DPMF  & 6.3687  $\pm$0.0081   & 6.5986 $\pm$ 0.0241 & 2.0512   $\pm$ 0.0021   & 2.1211  $\pm$ 0.0074   \\
    \hline
    \end{tabular}%
 \label{default}
\end{table*}%

\subsection{Experimental Results}

\subsubsection{Recommendation accuracy.}
To compare the effectiveness of these four methods, we evaluate their recommendation accuracy under the default privacy specification setting. The results are shown in Table \ref{default}. First, we can observe that non-personalized DPMF performs much worse than personalized HDPMF and PDPMF. This demonstrates that it is crucial to provide personalized privacy protection for different ratings, or we will face an intolerable utility loss. Second, our HDPMF has a significantly smaller average MSE and MAE than the SOTA PDPMF on all datasets, which verifies the superiority of our HDPMF framework. Third, when dimension $K=10$, HDPMF outperforms PDPMF by $5.97\%$, $5.00\%$, $9.61\%$, $8.15\%$  relatively on MSE in four datasets respectively. Besides, the relative gaps between them are $1.75\%, 2.40\%, 15.44\%, 1.98\%$ on MSE respectively when dimension $K=5$. Except on Douban dataset, the relative gap between HDPMF and SOTA PDPMF is apparently larger when dimension $K=10$ than dimension $K=5$. Thus, we can conclude that HDPMF achieves a better trade-off between privacy and accuracy than PDPMF, especially in high dimensional MF.

\begin{figure*}[htb!]
    \centering
    \includegraphics[scale=0.21]{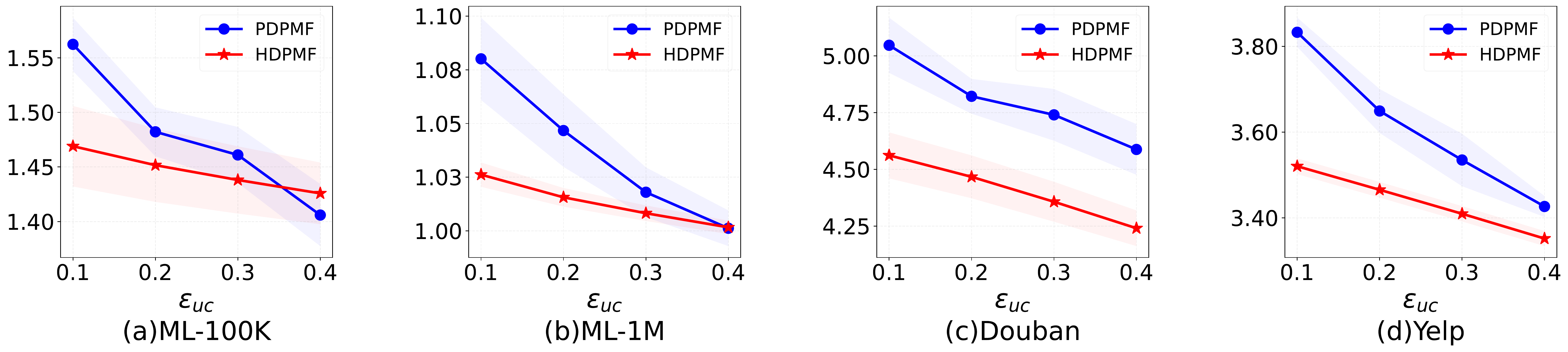}
    \caption{Average and standard deviation MSE of PDPMF and HDPMF with different $\epsilon_{uc}$ when model dimension $K=10$. DPMF performs much worse than PDPMF and HDPMF, the performance of MF stays the same as Table \ref{default}, so DPMF and MF are not shown for conciseness.}
    \label{eps_mse_10}
\end{figure*}

\subsubsection{Impact of privacy specification}
We also conduct various experiments by changing privacy specification parameter $\epsilon_{uc}$ in $\{0.1,0.2,0.3,0.4\}$ and $f_{uc}$ in $\{0.1,0.2,0.3,0.4,0.54\}$. The other privacy specification parameters stay the same as their default values in Table \ref{privacy} when varying $\epsilon_{uc}$ or $f_{uc}$. 

When model dimension $K=10$, the average MSE with different $\epsilon_{uc}$ is presented in Figure \ref{eps_mse_10}. And the situation is similar when $K=5$, so we hide it due to space limitation. As non-personalized DPMF performs much worse than personalized HDPMF or PDPMF in all experiments and the performance of non-private MF will not be affected by varying privacy parameter $\epsilon_{uc}$, it stays the same as the one in Table \ref{default}, so these two methods are not presented in Figure \ref{eps_mse_10} for conciseness. Also, due to space limitation and the same trend as MSE, we do not show the similar results in terms of MAE. From Figure \ref{eps_mse_10}, we can observe that HDPMF defeats PDPMF in most cases, especially when $\epsilon_{uc}$ is small (PDPMF performs slightly better only in several cases when $\epsilon_{uc}$ is relatively large), indicating that our HDPMF approach performs better than PDPMF when privacy guarantee is strong. 

\begin{figure*}[htb!]
    \centering
    \includegraphics[scale=0.21]{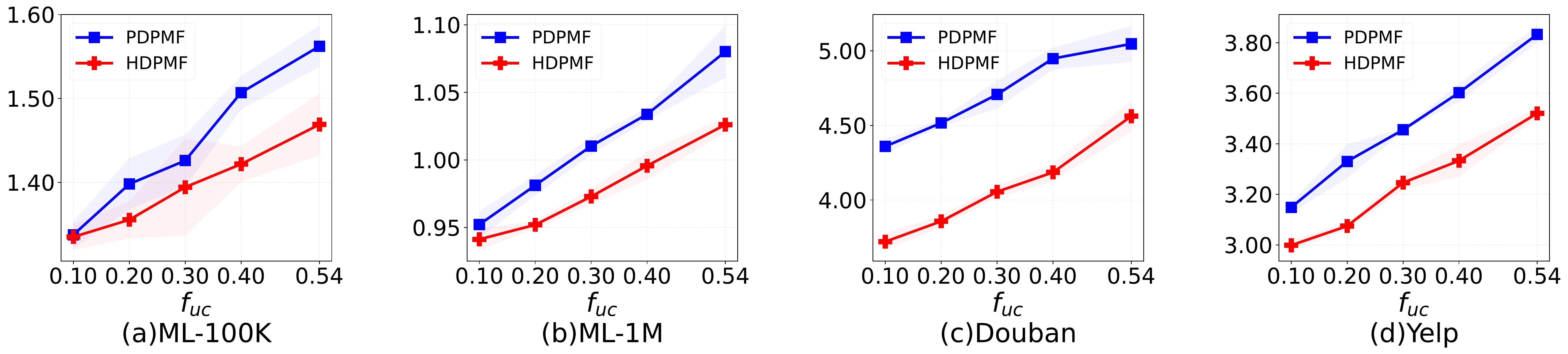}
    \caption{Average and standard deviation MSE of PDPMF and HDPMF with different $f_{uc}$ when model dimension $K=10$. DPMF performs much worse than PDPMF and HDPMF, the performance of MF stays the same as Table \ref{default}, so DPMF and MF are not shown for conciseness.}
    \label{fuc_mse_10}
\end{figure*}

The results with different $f_{uc}$ when model dimension $K=10$ are shown in Figure \ref{fuc_mse_10}. It should be noted that a larger $f_{uc}$ means that there are more conservative users so ratings have overall stronger privacy guarantees. From Figure \ref{fuc_mse_10}, we can find that HDPMF always outperforms PDPMF on average MSE with different $f_{uc}$. And the gap between HDPMF and PDPMF tends to become larger with a higher $f_{uc}$ in most cases. It further confirms the superiority of HDPMF over SOTA PDPMF in \emph{strong privacy guarantee scenario}.

\subsubsection{Impact of sparsity}
We then conduct experiments to test the impact of data sparsity on the performance of different methods. We change the sparsity of all datasets through sampling $\{20,40,60,80,100\}$ percentage of training ratings user by user. Because some users may have less than $10$ ratings after sampling, we use leave-one-out evaluation for these experiments. As for the privacy specification parameters, all of them stay the same as their default values in Table \ref{privacy}.

\begin{figure*}[htb!]
    \centering
    \includegraphics[scale=0.21]{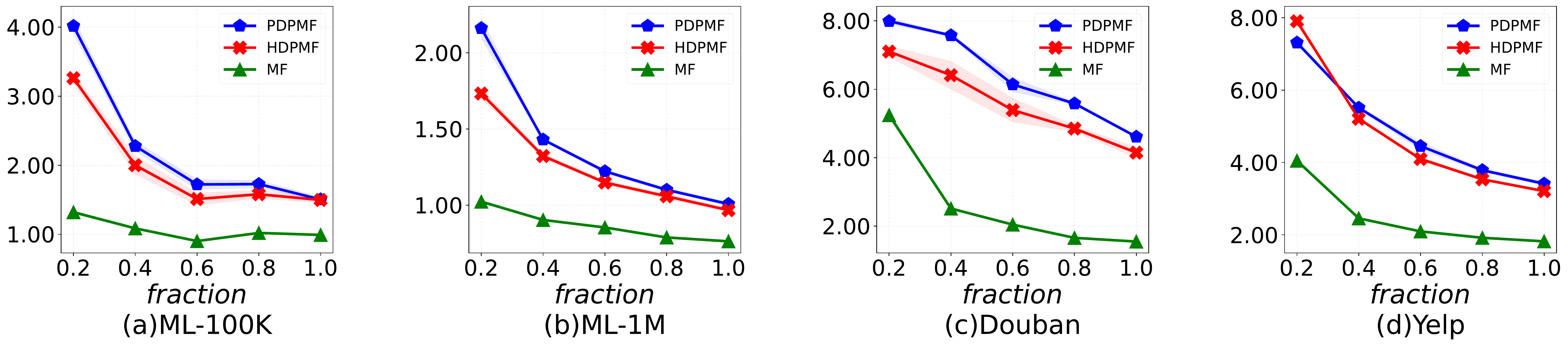}
    \caption{Average and standard deviation MSE of PDPMF and HDPMF with different dataset sparsity parameter $fraction$ when model dimension $K=10$. A smaller $fraction$ means a sparser training dataset. DPMF performs much worse than the other three methods, so it is not shown.}
    \label{sp_mse_10}
\end{figure*}

When model dimension $K=10$, the results of changing sparsity experiments are shown in Figure \ref{sp_mse_10}. Also, non-personalized DPMF is not presented because of much worse performance than personalized methods. And the performance of non-private MF will be affected by dataset sparsity, so its performance is shown. In Figure \ref{sp_mse_10}, it is obvious that the performance of all the three methods degrades when dataset becomes sparser, because the user preferences are harder to capture with less ratings. Besides, except very few cases on Yelp, HDPMF always outperforms PDPMF with different sparsity, and their performance gap tends to enlarge when the dataset becomes sparser. It further demonstrates the superiority of HDPMF over PDPMF, especially in \emph{sparse dataset scenario}.

\begin{figure}[htb!]
    \centering
    \includegraphics[scale=0.24]{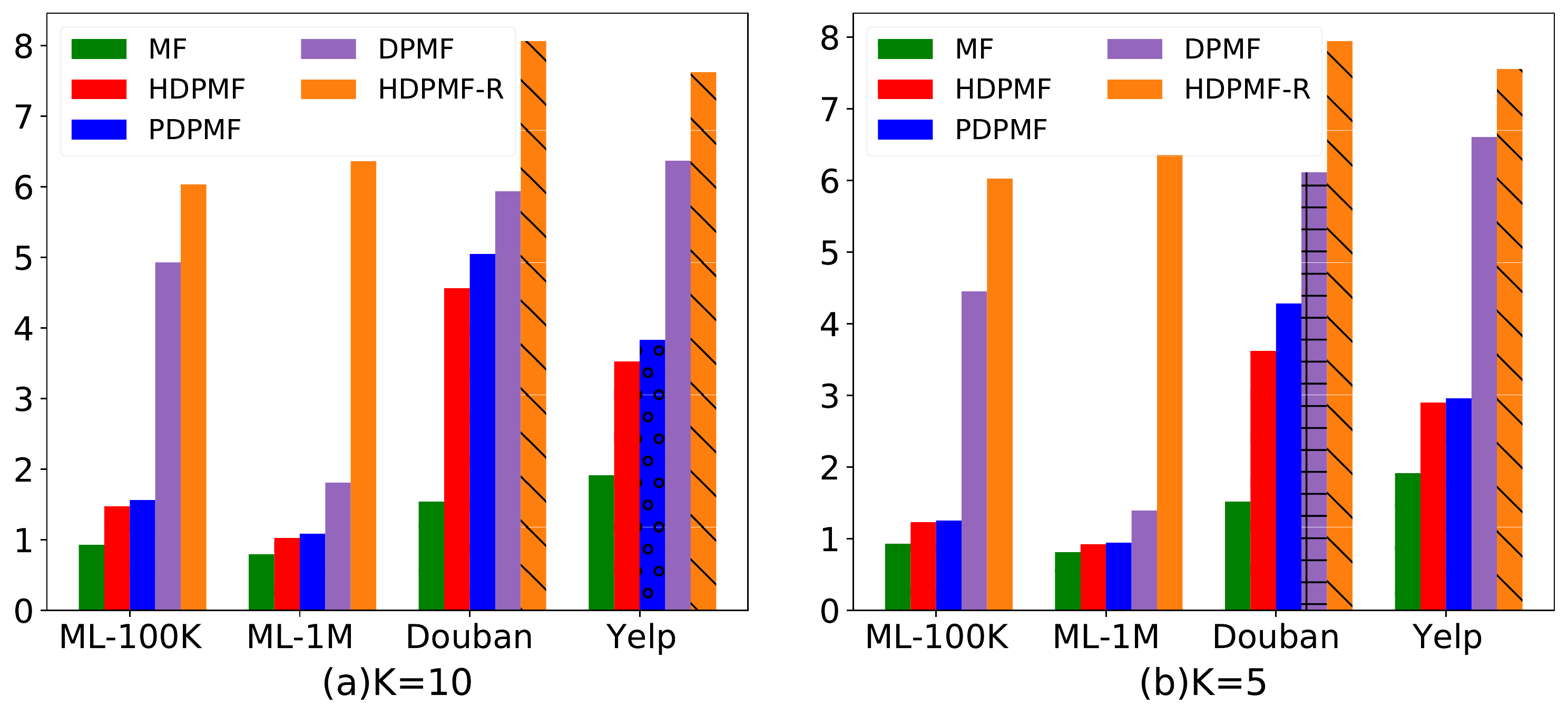}
    \caption{Average MSE of HDPMF, HDPMF-R and other baselines under default privacy specification setting. (a) shows results when model dimension $K=10$; (b) shows results when model dimension $K=5$. }
    \label{ablation}
\end{figure}

\subsubsection{Ablation study}
Our HDPMF framework utilizes an innovative rescaling scheme to tackle the abnormal scale problem \cite{jorgensen2015conservative} of original stretching mechanism. To verify its effectiveness, we conduct ablation study under default privacy specification setting. We derive the variant of HDPMF by removing rescaling scheme in the prediction stage and denotes it as "HDPMF-R". The performance of this variant and other baselines are presented in Figure \ref{ablation}, where the left subfigure shows the results when model dimension $K=10$ and the right subfigure shows the results when $K=5$. According to the performance of HDPMF and HDPMF-R, it is easy to know that removing rescaling scheme causes dramatic increase on average MSE. HDPMF-R even performs worse than the non-personalized DPMF method. A possible explanation is that in HDPMF-R, 
it uses distorted embeddings learned from stretched rating matrix to make predictions, thus the predicted ratings will have abnormal scale and bring huge error. It indicates that rescaling scheme is indispensable for getting normal scale rating predictions and achieving satisfying recommendation performance.

\section{Conclusion and Future Work}
To achieve personalized differential privacy in untrusted recommender scenario, we present the design and evaluation of a heterogeneously differentially private matrix factorization framework, denoted as HDPMF. It can satisfy personalized privacy requirements of different ratings through stretching mechanism. And a novel rescaling scheme is designed to achieve better trade-off between privacy and accuracy. Theoretical analysis and empirical evaluation demonstrate the superiority of our proposed framework.

Several promising directions can be further investigated. First, instead of point-wise model, we consider to extend our framework to pair-wise one \cite{DBLP:conf/uai/RendleFGS09}. Second, we are interested in extending our framework to deep neural networks such as NCF \cite{he2017neural}. 




\section*{Proof of Theorem 2}
\begin{proof}
Let $f(\boldsymbol{R})= \frac{\partial \mathcal{L}}{\partial \boldsymbol{v}_{j}} =2\sum_{i=1}^{n} \boldsymbol{I}_{i j}\left(\boldsymbol{u}_{i}^{T} \boldsymbol{v}_{j}-\boldsymbol{R}_{i j}\right) \boldsymbol{u}_{i} +2 \lambda \boldsymbol{v}_{j}$. $\boldsymbol{R},\boldsymbol{R}^{'}$ are neighboring datasets that differ only on the rating $\boldsymbol{R}_{pq}$. We flatten $\boldsymbol{R}$ into 
the input vector $\vec{d}$ and flatten $\boldsymbol{W}$ into the privacy weight vector $\vec{v}$, where $\boldsymbol{R}_{ij}=\vec{d}_k, \boldsymbol{W}_{ij}=v_k, k=m(i-1)+j$. And we take $T(\vec{v})=\operatorname{diag}(\vec{v})$, then $R(f,\vec{v})(\vec{d})=f(T(\vec{v}) \cdot \vec{d})=f(\operatorname{diag}(\vec{v})\cdot \vec{d})=f(\boldsymbol{W} \odot \boldsymbol{R})$, so the stretching mechanism can be formulated as follows.
\begin{equation*}
\begin{split}
    \mathcal{SM}(\vec{d},\vec{v},\epsilon)&=R(f,\vec{v})(\vec{d})+\boldsymbol{x}_j=f(\boldsymbol{W} \odot \boldsymbol{R})+\boldsymbol{x}_j=\frac{\partial \widetilde{\mathcal{L}}}{\partial \boldsymbol{v}_{j}} \\
    &=2\sum_{i=1}^{n} \boldsymbol{I}_{i j}(\boldsymbol{u}_{i}^{T} \boldsymbol{v}_{j}-\boldsymbol{W}_{ij}\boldsymbol{R}_{i j}) \boldsymbol{u}_{i} +\boldsymbol{x}_j+2 \lambda \boldsymbol{v}_{j}
\end{split}
\end{equation*}
where $\boldsymbol{x}_j \sim Laplace(2\sqrt{K}\Delta/\epsilon)$ denotes noise vector and $\odot$ is element-wise product.

From the neighboring datasets $\boldsymbol{R},\boldsymbol{R}^{'}$, we obtain the same derived $\boldsymbol{V}$. Since the derived $\boldsymbol{V}$ are optimized through gradient descent, we then have $\frac{\partial \widetilde{\mathcal{L}} (\boldsymbol{R})}{\partial \boldsymbol{v}_{j}}=\frac{\partial \widetilde{\mathcal{L}} (\boldsymbol{R}^{'})}{\partial \boldsymbol{v}_{j}}$. So we can know,
\begin{equation*}
\begin{split}
    &2\sum_{i=1}^{n} \boldsymbol{I}_{i j}(\boldsymbol{u}_{i}^{T} \boldsymbol{v}_{j}-\boldsymbol{W}_{ij}\boldsymbol{R}_{i j}) \boldsymbol{u}_{i} +\boldsymbol{x}_j \\
    &=2\sum_{i=1}^{n} \boldsymbol{I}_{i j}(\boldsymbol{u}_{i}^{T} \boldsymbol{v}_{j}-\boldsymbol{W}_{ij}\boldsymbol{R}^{'}_{i j}) \boldsymbol{u}_{i} +\boldsymbol{x}^{'}_j
\end{split}
\label{eq3}
\end{equation*}

As $\boldsymbol{R},\boldsymbol{R}^{'}$ differs only on the rating $\boldsymbol{R}_{pq}$ (or $\boldsymbol{R}^{'}_{pq}$), and we let $l=m(p-1)+q$, thus $\boldsymbol{W}_{pq}=v_l$, then we can get $\boldsymbol{x}_j-\boldsymbol{x}^{'}_j = 2\boldsymbol{W}_{pq}(\boldsymbol{R}_{pq}-\boldsymbol{R}^{'}_{pq}) \boldsymbol{u}_p$.


Considering $|\boldsymbol{R}_{pq}-\boldsymbol{R}^{'}_{pq}| \leq \Delta$ and $||\boldsymbol{u}_p||_2 \leq 1$, we know $||\boldsymbol{x}_j-\boldsymbol{x}^{'}_j||_2 \leq 2\boldsymbol{W}_{pq} \Delta=2 v_l \Delta$. We then formulate the probability that we get the same derived $\boldsymbol{V}$ with the neighboring datasets $\boldsymbol{R}$ and $\boldsymbol{R}^{'}$. For each vector $\boldsymbol{v}_j$  of $\boldsymbol{V}$,

\begin{equation*}
\begin{aligned}
&\frac{\operatorname{Pr}[\boldsymbol{v}_j|\boldsymbol{R}]}{\operatorname{Pr}[\boldsymbol{v}_j|\boldsymbol{R}^{'}]} =\frac{\prod_{u \in\{1,2, \ldots, K\}} \operatorname{Pr}\left(\boldsymbol{x}_{ju}\right)}{\prod_{u \in\{1,2, \ldots, K\}} \operatorname{Pr}(\boldsymbol{x}^{'}_{ju})} \\
&=\exp({-\frac{\epsilon \sum_{u}\left|\boldsymbol{x}_{ju}\right|}{2 \Delta \sqrt{K}}}) /\exp({-\frac{\epsilon \sum_{u}|\boldsymbol{x}^{'}_{ju}|}{2 \Delta \sqrt{K}}}) \\
&=\exp({\frac{\epsilon \sum_{u}(|\boldsymbol{x}^{'}_{ju}|-|\boldsymbol{x}_{ju}|)}{2 \Delta \sqrt{K}}}) \leq \exp({\frac{\epsilon \sum_{u}|\boldsymbol{x}_{ju}-\boldsymbol{x}^{'}_{ju}|}{2 \Delta \sqrt{K}}})  \\
& = \exp({\frac{\epsilon ||\boldsymbol{x}_j-\boldsymbol{x}^{'}_j||_1}{2 \Delta \sqrt{K}}}) \leq \exp({\frac{\epsilon \sqrt{K}||\boldsymbol{x}_j-\boldsymbol{x}^{'}_j||_2}{2 \Delta \sqrt{K}}}) \\
&\leq \exp({\frac{\epsilon \sqrt{K} 2 v_l \Delta}{2 \Delta \sqrt{K}}}) = \exp(\epsilon v_l)
\end{aligned}
\end{equation*}

So, we conclude that the derived item latent  matrix $\boldsymbol{V}$  satisfies heterogeneous $(\epsilon, \vec{v})$-differential privacy.
\end{proof}

\section*{Acknowledgment}
This work is supported by Shanghai Rising-Star Program (Grant No. 23QA1403100), Natural Science Foundation of Shanghai (Grant No. 21ZR1421900), National Natural Science Foundation of China (Grant No. 72192832), Graduate Innovation Fund of Shanghai University of Finance and Economics (Grant No.CXJJ-2022-366) and the Program for Innovative Research Team of Shanghai University of Finance and Economics.



\bibliographystyle{unsrt}
\bibliography{reference}

\end{document}